\pdfoutput=1

\documentclass[11pt]{article}

\usepackage[]{emnlp2021}

\usepackage{times}
\usepackage{latexsym}

\usepackage[T1]{fontenc}

\usepackage[utf8]{inputenc}

\usepackage{microtype}
\usepackage{multirow}
\usepackage{booktabs}
\usepackage{graphicx}
\title{Decomposing Complex Questions Makes Multi-Hop QA Easier and More Interpretable}

\author{Ruiliu Fu\textsuperscript{1,2} \and Han Wang\textsuperscript{1,2} \and Xuejun Zhang\textsuperscript{1,2} \and Jun Zhou\textsuperscript{1} \and Yonghong Yan\textsuperscript{1} \\
  \textsuperscript{1}Institute of Acoustics, Chinese Academy of Sciences \\
  \textsuperscript{2}University of Chinese Academy of Sciences \\
  \texttt{$\{$furuiliu,wanghan,zhangxuejun,zhoujun,yanyonghong$\}$@hccl.ioa.ac.cn} }

\begin{document}
\maketitle
\begin{abstract}
 Multi-hop QA requires the machine to answer complex questions through finding multiple clues and reasoning, and provide explanatory evidence to demonstrate the machine’s reasoning process. We propose \textbf{R}elation \textbf{E}xtractor-\textbf{R}eader and \textbf{C}omparator (RERC), a three-stage framework based on complex question decomposition. The Relation Extractor decomposes the complex question, and then the Reader answers the sub-questions in turn, and finally the Comparator performs numerical comparison and summarizes all to get the final answer, where the entire process itself constitutes a complete reasoning evidence path. In the 2WikiMultiHopQA dataset, our RERC model has achieved the state-of-the-art performance, with a winning joint F1 score of 53.58 on the leaderboard. All indicators of our RERC are close to human performance, with only 1.95 behind the human level in F1 score of support fact. At the same time, the evidence path provided by our RERC framework has excellent readability and faithfulness.
\end{abstract}

\section{Introduction}

Multi-hop QA is an important and challenging task in natural language processing (NLP), which requires complex reasoning over several paragraphs to reach the final answer and explanatory evidence to demonstrate the reasoning process. Many high-quality multi-hop QA datasets have been introduced recently, such as HotpotQA~\citep{Yang:18}, ComplexWebQuestions~\citep{Talmor:18}, QAngaroo WikiHop~\citep{Welbl:18}, $R^4C$~\citep{Inoue:20}, 2WikiMultiHopQA~\citep{Ho:21}, etc.

These high-quality multi-hop QA datasets promote many multi-hop QA models~\citep{Song:18,Ding:19,Xiao:19,Nishida:19,Tu:19,Cao:19}, most of which are end-to-end models based on graph structure or graph neural network~\citep{Veli:18}. Although these works have good performances in many tasks, they also have some limitations to address. First of all, the internal reasoning mechanism of previous end-to-end QA models is a black-box, which usually use an additional discriminator to judge whether a sentence is a clue sentence, such as DFGN~\citep{Xiao:19}. There is no evidence to show that such additional discriminators are strongly correlated with the reasoning results of the end-to-end model, which means not faithful. Secondly, although graph structure is helpful to multi-hop reasoning in theory, but recent work~\citep{Shao:20} shows that the existing graph neural network is only a special attention mechanism~\citep{Bahdanau:14}, and it's not necessary for multi-hop QA, with the experiments that better results can be achieved by using only transformer network instead of graph neural network, as long as the same additional adjacency matrix information is provided.

We observed that human reasoning about complex questions is not accomplished overnight and it's usually divided into the steps of question decomposition, answering sub-questions, summarizing and comparing. For example, for the complex question, "whose candidate will get more votes in the 2020 U.S. election, Democrats and Republicans?" People will not think about the whole question, but firstly decompose the complex question. Realizing that the subject of the question is "Democrats and Republicans", and the question is about "candidates" and "number of votes", people can answer those sub-questions progressively -- "who is the Democratic candidate?" and "how many votes does \textbf{ANS} get?" The same thinking process was performed for another question subject, "Republican Party". Finally, the two votes were compared to obtain the answer to the entire complex question.

Inspired by the way humans answer complex multi-hop questions, in this work we abandoned the end-to-end model structure, but imitated the human reasoning mechanism to propose a three-stage Relation Extractor-Reader and Comparator (RERC) model\footnote{Our source code is available in \url{https://github.com/furuiliu/RERC}.}. We first build a Relation Extractor, which can automatically extract the subject and key relations of the question from the complex unstructured textual representation. For the Relation Extractor, we use two different structures, one is classification-type (CRERC), where the evidence relation information in the dataset is used as prior knowledge, and the question text is mapped to question relations through the classifier; the other is span-type (SRERC), where the type of question relations is unrestricted, and the Relation Extractor can automatically extract multiple corresponding spans from the question text as question relations. Next, we use the advanced ALBERT model~\citep{Lan:20} as the Reader, which reads the corresponding paragraphs and answer each sub-question composed of the subject and relations of the question in turn. Finally, for comparison type questions, our Comparator module compares the magnitude of each subject's final answer, and then get the entire answer.

Our contributions are summarized as follows:
\begin{itemize}
\setlength{\itemsep}{0pt}
\setlength{\parsep}{0pt}
\setlength{\parskip}{0pt}
\item[$\bullet$] We propose a novel RERC model for multi-hop text-based QA and evidence path search tasks.
\item[$\bullet$] We propose a Query-aware Entity Tree Paragraph Screening (QETPS) method  to filter valid paragraphs from a large number of documents before Reader module, which is more efficiently than previous paragraph selecting methods.
\item[$\bullet$] We provide an experimental study on a public multi-hop dataset (2WikiMultiHopQA) to demonstrate that our proposed RERC model has the state-of-the-art performance in both answering multi-hop questions and extracting evidence at the same time.
\end{itemize}

\begin{figure*}[!htb]
\centering
\includegraphics[width=16cm]{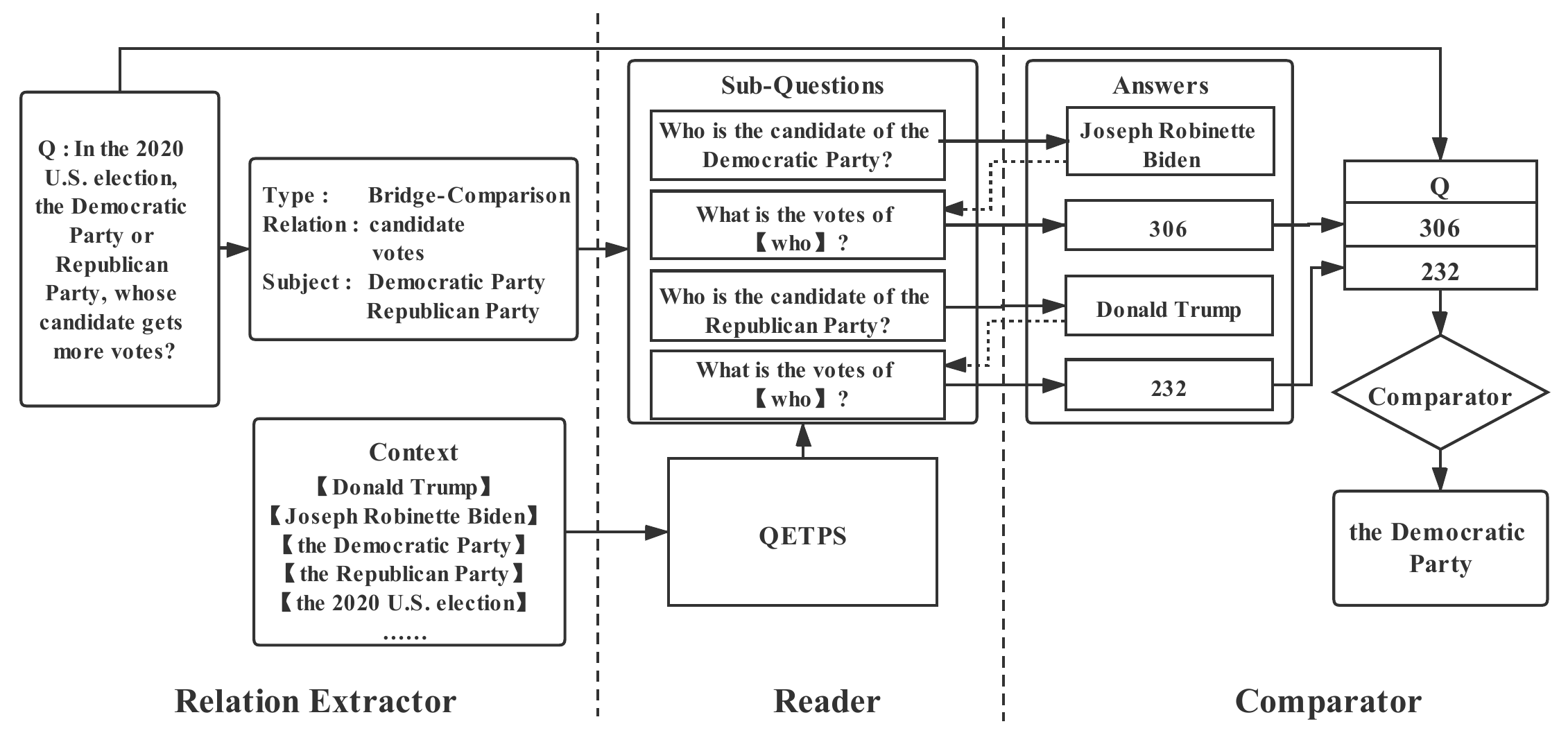}
\caption{Relation Extractor-Reader and Comparator (RERC) model}
\label{model-fig}
\end{figure*}

\section{Related work}

\subsection{Multi-hop QA research}
Initially, researchers still has been using the previous ideas in single-hop reading comprehension, focusing on the query-document co-inference attention method~\citep{Dhingra:18,Zhong:19,Cao:19}. Until Ding et al.~\shortcite{Ding:19} cleverly applied the graph neural network to the multi-hop QA task, and achieved excellent performance improvement, then other models such as DFGN~\citep{Xiao:19} were successively proposed to integrate graph structure into multi-hop QA tasks.

However, recently these end-to-end methods in multi-hop QA tasks seem to have fallen into a bottleneck that there is still a huge gap from human level. Besides, the internal reasoning process of these end-to-end multi-hop QA models is not clear, and the generated explanations are not faithful enough. Our proposed Relation Extractor-Reader and Comparator (RERC) model adopts the idea of decomposing complex questions. It decomposes complex multi-hop QA tasks into multiple single-hop reading comprehension subtasks, and transforms complex tasks into simple tasks that we have solved. In this way, the RERC model has successfully avoided the dilemmas of unclear internal mechanism and unfaithful interpretation caused by the separation of interpretation and reasoning, which the above-mentioned existing end-to-end models have faced.

\subsection{Complex question decomposition}
Complex question decomposition is also an important task in NLP area, which is closely related to multi-hop QA task. For example, the DecompRC model~\citep{Min:19} regarded the complex question decomposition as a span extraction task, and used a supervised model to decompose the complex question into multiple spans to solve the multi-hop QA task. However, this method of using question spans as sub-questions is only suitable for specific Compositional-type complex questions. Not all complex questions can be decomposed into sub-questions by question fragments. ONUS~\citep{Perez:20} adopted an unsupervised method, using the characteristics of HotpotQA~\citep{Yang:18} multi-hop QA dataset and SQuAD~\citep{Rajpurkar:16,Rajpurkar:18} single-hop reading comprehension dataset which are both based on Wikipedia document, and used similar matching to construct some pseudo-data from complex questions to simple questions, and then trained an unsupervised sequence-to-sequence (seq2seq) model~\citep{Artetxe:18} to generate sub-questions. The method relies on the homology characteristics of the two datasets HotpotQA~\citep{Yang:18} and SQuAD~\citep{Rajpurkar:16,Rajpurkar:18}, which is more restrictive.

In this work, we propose that the complex question decomposition model used for multi-hop QA does not need to generate complete sentence-type sub-questions
 and most complex questions cannot be directly divided into complete sub-questions. We only need to extract the key question subjects and question relations, and then construct them through templates, which not only reduce the difficulty of decomposition of complex questions, but also apply to the decomposition of complex questions of any form.

\section{Proposed method}
\label{method}

In this section, we will introduce in detail our proposed Relation Extractor-Reader and Comparator (RERC) model. This is a three-stage multi-hop QA model consisting of three parts: Relation Extractor, Reader and Comparator. The working principle of the whole framework is shown in figure ~\ref{model-fig}. Given the question $q$ and the context set $C=\left\{c_{i}\right\}$, firstly pass the question $q$ to the Relation Extractor to obtain the question subjects set $E=\left\{e_{i}\right\}$ and the question relations set $R=\left\{r_{i}\right\}$, and then construct the sub-questions set $ SQ=\left\{sq_{i}\right\}$, and then the Reader reads the searched context and answers each sub-question to obtain the answer set $A=\left\{a_{i}\right\}$, and finaly the Comparator obtains the final answer $ANS$ through numerical comparison and summary analysis.

\subsection{Relation Extractor}
In this work, we have experimented with two Relation Extractors, named Classification-type Relation Extractor (CRE) and Span-type Relation Extractor (SRE). The difference between the two Relation Extractors is whether to use the evidence relation information in the dataset, so they are only distinguished in the output layer.

The Classification-type Relation Extractor (CRE) is firstly introduced, which uses an advanced text classifier structure. We first use the advanced large-scale pre-training language model BERT~\citep{Devlin:18} to encode the question $q$ to obtain a question encoding representation $Q_{Embed}$ with rich semantic information.

Next, we need to perform self-interaction calculations on the entire sentence and find the relationship between the words in the sentence through the self-attention mechanism, so as to find the key information corresponding to the subject and the relations of the question. We use the Transformer network based on the self-attention mechanism ~\citep{Vaswani:17} as our interaction layer, by encoding the question representation $Q_{Embed}$ obtained above to express self-interaction, and then get the question self-interaction representation $Q_{Inter}$ and the question pooling representation $Q_{Pooled}$ after MaxMeanPooler pooling operation.

\begin{equation}
Q_{Inter}=Transformer(Q_{Embed})
\label{Q_Inter}
\end{equation}

In order to make specific reasoning for different question types (Compositional, Inference, Comparison and Bridge-Comparison), we need to determine the question type $Type$ first. Therefore, we deploy a linear classification layer $TypeLinear$ to calculate the probability of the four question types $Q_{Type}$ and category prediction $T=argmax(Q_{Type})$:

\begin{equation}
Q_{Type}=TypeLinear(Q_{Pooled})
\label{Q_Type}
\end{equation}

For the four different question types, we use four independent relation classifiers $\left\{RelationLinear_{i}\right\}$, and then use the category-aware fusion mechanism to fuse the results of the four relation classifiers to get the final relation prediction result $R$:

\begin{equation}
R=argmax(Q_{Type}\cdot r)
\label{R}
\end{equation}

where $r=\left [r_1,r_2,r_3,r_4\right ]$, $r_{i}=RelationLinear_{i}(Q_{Pooled}),\: i=1,2,3,4$.

Besides, we also predict the question subject entity, which is a sequence span extraction task. We choose a pointer network ~\citep{Vinyals:15} EntityPointer to perform this task:

\begin{equation}
E=EntityPointer(Q_{Inter})
\label{Q_E}
\end{equation}

The loss function of the Relation Extractor is designed as $loss = loss_R +\alpha \cdot loss_T+\beta \cdot loss_E$, where $loss_R$, $loss_T$ and $loss_E$ represent the prediction loss of the question relation, question type, and the question subject respectively.

Above is the detailed structure of the entire Classification-type Relation Extractor (CRE). However, the CRE model must be required to limit the known relation categories, which greatly limits its versatility. Therefore, we additionally propose a Span-type Relation Extractor (SRE) to replace relation category prediction with relation span extraction. We also use four pointer networks $\left\{RelationPointer_{i}\right\}$ to perform relation span extraction, and then perform category-aware fusion. The whole process is basically the same as the CRE model, so we don’t repeat it here.

After obtaining the prediction results of the subjects and the relations of the question, they are spliced together to form the sub-questions set $SQ=\left\{sq_{i}\right\}$ which are sent to the next Reader module.

\begin{equation}
SQ=\left\{e_{i}|r_{j}\: \: \forall e_{i}\in E,r_{j}\in R\right\}
\label{SQ}
\end{equation}

\subsection{Sub-Question Reader}
\label{singreader}
Before reading comprehension, we need to sort or filter all the paragraphs, because our model has a limited ability to process long-sequence texts, and the total length of the context in the task greatly exceeds this limit, which is also common in practical applications, and most of context is useless to answer the sub-questions. We propose a Query-aware Entity Tree Paragraph Screening (QETPS) method.

Through careful observation, we find that every hop in the multi-hop QA dataset needs to pass through the entity (person, organization, location, etc.) as a transfer, which is also in line with our common sense of life. Therefore, we can build an entity tree through the interdependence between entities to make each paragraph sorted according to priority.

Specifically, we first locate all entities in the question sentence and use these entities as the root nodes of the entity tree. Then we look for the paragraphs where these root entities appear, and associate those entities that appear in the same sentence with root entities as the child nodes. Then we start from these child nodes and repeat the above process until no new child nodes can be added to the tree, at this time our entity tree is formed. In order to prevent the influence of interfering paragraphs, we have added a query-aware regulation mechanism that only the child nodes in the corresponding sentence of the query can be added. At the same time, in order to ensure the effectiveness of the method, we did not use exact matching(EM) when searching for the corresponding entities or relations. Instead, we used the F1 value calculated by the longest common subsequence length as the similarity, by setting threshold to determine whether it appears.

After constructing the entity tree, we believe that the answer for the $i$th-hop sub-question is most likely to exist in the paragraph associated with the node at the $i$th level of the entity tree (the root node is the $0$th level). So we successively obtain the filtered paragraph representation $C_{QETPS}$ through adding paragraphs corresponding to nodes according to the distance in the tree.

Next, we use the advanced AlbertForQuestionAnswering model~\citep{Lan:20} as Reader to answer each sub-question, and get the answer set $A=\left\{a_{i}\right\}$:

\begin{equation}
a_{i}=Reader(sq_{i}\: |\: C_{QETPS}^i),\: i=1,2,3,...
\label{SRC}
\end{equation}

\subsection{Comparator}
After getting the answers to all sub-questions, we need to summarize these answers to get the final answer, which also depends on the question type we get in the Relation Extractor. For Compositional-type and Inference-type questions, we only need to output the answer of the last sub-question. So we should focus on Comparison-type and Bridge-Comparison-type questions.

We trained a Comparator that can compare various types of quantitative relationship problems universally. We splice the question text description and the two objects to be compared, and send them to the quantity relationship Comparator to get the comparison result $A_{Compare}$:

\begin{equation}
A_{Compare}=Comparator(q\: |\: \hat{a_1}\: |\: \hat{a_2})
\label{Compare}
\end{equation}

where $\hat{a_1}$ and $\hat{a_2}$ respectively represent the last sub-answer corresponding to the two question subjects, and the four states of the comparison result $A_{Compare}\in R^{4}$ are respectively represents -- "0: not equal, 1: equal, 2: the first option meets, 3: the last option meets". 

\begin{table*}[htbp]
  \centering
  \resizebox{\textwidth}{!}{
    \begin{tabular}{clcccccccc}
    \toprule
    \multirow{2}[3]{*}{} & \multicolumn{1}{c}{\multirow{2}[3]{*}{\textbf{Model}}} & \multicolumn{2}{c}{\textbf{Answer}} & \multicolumn{2}{c}{\textbf{Sp fact}} & \multicolumn{2}{c}{\textbf{Evidence}} & \multicolumn{2}{c}{\textbf{Joint}} \\
\cmidrule{3-10}          &       & EM    & F1    & EM    & F1    & EM    & F1    & EM    & F1 \\
    \midrule
    \multirow{9}[2]{*}{Dev} & Ho et al.~\shortcite{Ho:21} & 35.30  & 42.45  & 23.85  & 64.31  & 1.08  & 14.77  & 0.37  & 5.03  \\
          & Yang et al.~\shortcite{Yang:18} & 34.14  & 40.95  & 26.47  & 66.94  & -     & -     & -     & - \\
          & *DFGN~\citep{ Xiao:19} & 30.87  & 38.49  & 17.06  & 57.79  & -     & -     & -     & - \\
          & *QFE~\citep{Nishida:19} & 37.56  & 43.21  & 21.13  & 59.20  & -     & -     & -     & - \\
          & *QFE + Evidence Extractor & 38.30  & 44.22  & 34.62  & 72.18  & 6.62  & 33.68  & 3.57  & 13.53  \\
          & *DecompRC~\citep{Min:19} & 7.46  & 41.57  & 56.49  & 82.73  & -     & -     & -     & - \\
          & *DecompRC + Comparator & 39.94  & 61.46  & 68.45  & 85.54  & -     & -     & -     & - \\
          & CRERC & \textbf{71.56 } & \textbf{74.51 } & \textbf{86.00 } & \textbf{92.75 } & \textbf{55.88 } & \textbf{70.32 } & \textbf{50.59 } & \textbf{60.21 } \\
          & SRERC & 69.74  & 73.81  & 81.89  & 89.95  & 8.26  & 25.67  & 7.66  & 21.80  \\
    \midrule
    \multirow{3}[1]{*}{Test} & Ho et al.~\shortcite{Ho:21} & 36.53  & 43.93  & 24.99  & 65.26  & 1.07  & 14.94  & 0.35  & 5.41  \\
          & Human & 80.67  & 82.34  & 85.33  & 92.63  & 57.67  & 75.63  & 53.00  & 66.69  \\
          & CRERC & \textbf{69.58 } & \textbf{72.33 } & \textbf{82.86 } & \textbf{90.68 } & \textbf{54.86 } & \textbf{68.83 } & \textbf{49.80 } & \textbf{58.99 } \\
    \bottomrule
    \end{tabular}}%
  \caption{Results on the development set and the test set of 2WikiMultiHopQA dataset. The mark * means the models we reproduced according to the open source code and the settings in the original paper. The mark - means those models have no ability to extract the evidence result.}
  \label{main_result_table}%
\end{table*}%

\section{Experiment}
\subsection{Dataset}
We use 2WikiMultiHopQA dataset\footnote{The dataset benchmark platform located at \url{https://github.com/Alab-NII/2wikimultihop.}} newly proposed by Ho et al.~\shortcite{Ho:21} to implement the experiments. The 2WikiMultiHopQA dataset contains a total of 192,606 questions jointly constructed through the Wikipedia document set and the Wikidata knowledge base, all of which require multi-hop reasoning. The dataset follows the similar design of HotpotQA~\citep{Yang:18}, and the data are split into a training set (167454 questions), a development set (12576 questions) and a test set (12576 questions). All questions in development and test sets are hard multi-hop cases. At the same time, the 2WikiMultiHopQA dataset is also divided into four different question types, namely Compositional, Inference, Comparison and Bridge-comparison.

Compared with HotpotQA~\citep{Yang:18}, the 2WikiMultiHopQA dataset removes simple-level questions, increases the types of questions, and the length of the questions and the diversity of answer forms. In addition to following the setting of HotpotQA, Ho et al.\shortcite{Ho:21} also added the prediction task of the evidence path, which further tested the reasoning and interpretation capabilities of the multi-hop QA model.

The performance evaluation of 2WikiMultiHopQA dataset takes into account the evaluation of the answer, the supporting facts, and the evidence path, using two evaluation metrics: exact match (EM) and F1 score.

\subsection{Experimental Details}
The Relation Extractor-Reader and Comparator (RERC) model we proposed is divided into three independently trained modules: Relation Extractor, Reader and Comparator.

\textbf{Relation Extractor} uses pre-trained BERT-base model released by Devlin et al.~\citep{Devlin:18} with question length $l=128$, hidden layer size $d=768$. 

For the CRE model, we collect the relation labels in the given evidence path in the dataset as the classification category labels, a total of 35 categories; for the SRE model, we construct 1,000 samples according to the relation span in the question text through crowdsourcing to train the span extraction pointer network.

\textbf{Reader} uses the ALBERT-large model released by Lan et al.~\shortcite{Lan:20} with $l=512$ and $d=1024$, which has been shown advanced performance in the SQuAD 1.1/2.0 dataset~\citep{Rajpurkar:16,Rajpurkar:18}.

\textbf{Comparator} use the model structure similar to the CRERC model with $l=256$ and $d=768$. 

\textbf{During training}, we use the Adam optimizer in all three modules, set the $batch \: size$ to 32,16,32, and the learning rate of $2\times 10^{-5}$, $1\times 10^{-5}$, $2\times 10^{-5}$ separately. The learning rate for parameters in BERT warmup over the first $10\%$ steps, and then linearly decays to zero. The hyperparameter of the loss function in RE is set to $\alpha=\beta=1.0 $.

In addition, we also proposed the QETPS method described in the section ~\ref{singreader}. We use the Named Entity Recognition (NER) tool $Stanford \: corenlp \: toolkit$~\citep{Manning:14} to extract the corresponding named entities from all texts, and then use the threshold of $\sigma_1=0.8$ and $\sigma_2=0.65$ to match the entity nodes and question relation.

All experiments are based on four Tesla P100 GPUs. In order to determine the proposed method in each stage, we compared a variety of methods through experiments which are described at Appendix ~\ref{appendix_alter}.

\subsection{Baseline}
We will compare the performance of our RERC model and the previous works on the 2WikiMultiHopQA dataset ~\citep{Ho:21}.

\textbf{Ho et al.~\shortcite{Ho:21}} The strong baseline model released in the original 2WikiMultiHopQA paper~\citep{Ho:21}. It was based on the multi-hop model proposed by Yang et al.~\shortcite{Yang:18}, and added a new component to perform the evidence generation task.

\textbf{DFGN~\citep{Xiao:19}} The classic end-to-end multi-hop QA model based on graph neural network, originally working on HotpotQA~\citep{Yang:18} dataset. We reproduced the DFGN model by using the BERT-base pre-trained model~\citep{Devlin:18} under the source code and hyperparameter settings published by Yang et al.~\shortcite{Yang:18}.

\textbf{DecompRC~\citep{Min:19}} The classic multi-hop QA model that using question decomposition methods, originally working on HotpotQA~\citep{Yang:18} dataset. We reproduced the DecompRC model by using the same question decomposition method as Min et al.~\shortcite{Min:19} and the same Reader module as our RERC model, which is helpful to compare our method with the DecompRC model in question decomposition.

\textbf{QFE~\citep{Nishida:19}} The classic multi-hop QA model which was based on the multi-hop model proposed by Yang et al.~\shortcite{Yang:18}, and added a Query-Focused Extractor(QFE) module to extract the supporting sentences. We reproduced the QFE model following Nishida et al.~\shortcite{Nishida:19}.

\textbf{Human} Ho et al.~\shortcite{Ho:21} randomly selected 100 samples in the test set to evaluate human performance.

Next is the introduction of some variants,

\textbf{CRERC -wo QETPS} The CRERC model which does not use the QETPS method but add all paragraphs.

\textbf{CRERC -w PSBERT} The CRERC model which does not use the QETPS method but the paragraph selector of the BERT model applied in DFGN~\citep{Xiao:19}.

\textbf{DecompRC + Comparator} The variant of the \textbf{DecompRC} model of which the final answer is obtained through the Comparator module proposed in this work.

\textbf{QFE + Evidence Extractor} The variant of the \textbf{QFE} model which adds the same Evidence Extractor component as the original baseline model~\citep{Ho:21}.

\begin{table*}[htbp]
   \centering
   \resizebox{\textwidth}{!}{
     \begin{tabular}{cccccccccc}
     \toprule
     \multirow{3}[5]{*}{model} & \multicolumn{6}{c}{Relation Extractor} & \multicolumn{2}{c}{\multirow{2}[3]{*}{ Reader}} & \multirow{2}[3]{*}{Comparator} \\
\cmidrule{2-7} & \multicolumn{2}{c}{question subject} & \multicolumn{3}{c}{question relation} & \multicolumn{1}{l}{question type} & \multicolumn{ 2}{c}{} & \\
\cmidrule{2-10} & EM & F1 & Accuracy & EM & F1 & Accuracy & EM & F1 & Accuracy \\
     \midrule
     CRERC & 0.860 & 0.955 & 0.999 & - & - & 1.000 & 0.940 & 0.958 & 0.976 \\
 SRERC & 0.860 & 0.955 & - & 0.997 & 0.997 & 1.000 & 0.916 & 0.942 & 0.976 \\
     \bottomrule
     \end{tabular}}%
      \caption{Evaluation of each sub-module in RERC three-stage model. The accuracy of the question relation is only for the CRERC model, while the EM and F1 values only for SRERC model.}
   \label{sub_model_table}%
\end{table*}%

\begin{table*}[htbp]
  \centering
    \begin{tabular}{lcccccccc}
    \toprule
    \multicolumn{1}{c}{\multirow{2}[4]{*}{Type}} & \multicolumn{2}{c}{Answer} & \multicolumn{2}{c}{Sp fact} & \multicolumn{2}{c}{Evidence} & \multicolumn{2}{c}{Joint} \\
\cmidrule{2-9} & EM & F1 & EM & F1 & EM & F1 & EM & F1 \\
    \midrule
    Comparison & 72.96 & 73.22 & \textbf{96.22} & \textbf{98.20} & \textbf{87.80} & \textbf{93.68} & 67.11 & 69.71 \\
    Inference & 58.30 & 66.35 & 73.60 & 85.18 & 32.92 & 46.03 & 32.73 & 42.10 \\
    Compositional & 63.88 & 68.32 & 79.74 & 88.97 & 36.17 & 53.38 & 36.00 & 47.00 \\
    Bridge-Comparison & \textbf{92.11} & \textbf{92.31} & 93.60 & 98.19 & 71.07 & 90.41 & \textbf{70.16} & \textbf{85.06} \\
    \midrule
    All & 71.56 & 74.51 & 86.00 & 92.75 & 55.88 & 70.32 & 50.59 & 60.21 \\
    \bottomrule
    \end{tabular}%
  \caption{CRERC model performance under different question types}
  \label{Cmodel_ontype}%
\end{table*}%

\subsection{Results}
Table ~\ref{main_result_table} shows the evaluation result of our proposed Relation Extractor-Reader and Comparator (RERC) model on the development set and the test set of 2WikiMultiHopQA dataset~\citep{Ho:21}. Our proposed Classification-type Relation Extractor-Reader and Comparator (CRERC) model outperforms all competitors in the evaluation metrics of answer, support facts and evidences on the development set and the test set. Compared with human performance, our CRERC model is close to human performance in the evaluation metrics of support facts and evidences, with only 1.95 gap in F1 score of support facts. Although the objective indicators of the SRERC model for evidence are low, the evidence path generated by the SRERC model have better readability through human subjective observations, which we will describe in detail in the section ~\ref{people_test}.

In addition to the overall performance evaluation of the model, we also conducted a separate performance evaluation for each part of the three-stage modules. The specific evaluation results are shown in the table ~\ref{sub_model_table}, where the accuracy of the question relation is only for the CRERC model, and the EM and F1 values are only for SRERC model.

In the table ~\ref{sub_model_table}, we find that for the Relation Extractor module and the Comparator module, our proposed model has reached very high accuracy, which may be due to the fact that there are a few types of question relations and quantitative relationship comparison in the 2WikiMultiHopQA dataset. The performance of the Reader module has also reached such amazing accuracy as $EM=0.940$ and $F1=0.958$. Therefore, the future research of the question decomposition multi-hop QA model should focus on how to reduce the cumulative error of multiple hops and how to recognize and redress the errors of the previous reasoning steps when performing the next reasoning step. 

\section{Discussion}
\label{discussion}

\begin{table*}[htbp]
  \centering
    \begin{tabular}{lcccccccc}
    \toprule
    \multicolumn{1}{c}{\multirow{2}[3]{*}{Model}} & \multicolumn{2}{c}{Answer} & \multicolumn{2}{c}{Sp fact} & \multicolumn{2}{c}{Evidence} & \multicolumn{2}{c}{Joint} \\
\cmidrule{2-9} & EM & F1 & EM & F1 & EM & F1 & EM & F1 \\
    \midrule
    CRERC & \textbf{71.56} & \textbf{74.51} & \textbf{86.00} & \textbf{92.75} & \textbf{55.88} & \textbf{70.32} & \textbf{50.59} & \textbf{60.21} \\
    CRERC -wo QETPS & 37.13 & 38.79 & 20.89 & 54.34 & 6.63 & 16.72 & 0.07 & 2.27 \\
    CRERC -w PSBERT & 68.77 & 71.77 & 81.54 & 88.27 & 53.64 & 67.12 & 46.27 & 55.67 \\
    \bottomrule
    \end{tabular}%
  \caption{Results of Ablation experiment about QETPS method}
  \label{wo_qetps_table}%
\end{table*}%

\subsection{Impact of different question types}
To study the impact of different question types in the 2WikiMultiHopQA dataset, we perform some experiments to compare the CRERC model under each question type, where the results are shown in the table ~\ref{Cmodel_ontype}. We observed the best performance for our CRERC model in the Bridge-Comparison questions, which combine the Compositional-type and Comparison-type, and have the most number of hops and support facts to to be retrieved, and are designed to be the most challenging question type. We analyzed that it is due to our CRERC model's special method of decomposing complex questions based on relation extraction, which is not interfered by the expression of compound question types. Besides we find the question relation setting of Bridge-Comparison questions is relatively simple, and the sub-question is easier to answer, which offset the impact of more hops.

In general, the RERC model performs significantly better on Comparison-type and Bridge-Comparison-type than Compositional-type and Inference-type, which is due to that the Comparison-type and Bridge-Comparison-type questions have easier sub-questions, as compensation for additional comparison tasks, which can be accomplished greatly by our Comparator module.

\begin{table}[htbp]
  \centering
    \begin{tabular}{ll}
    \toprule
    Model & manual scoring \\
    \midrule
    CRERC & $4.03\pm 0.58$ \\
    SRERC & $4.22\pm 0.52$ \\
    \bottomrule
    \end{tabular}%
  \caption{Manual evaluation of evidence path}
  \label{people_score_table}%
\end{table}%

\subsection{Impact of QETPS}
Due to the length limitation of text the Reader module can process one time and the large number and long lengths of context in the dataset, we designed a Query-aware Entity Tree Paragraph Screening (QETPS) method to filter these paragraphs. In order to verify whether the QETPS method we introduced is effective, we executed ablation experiments to compare the performance changes after replacing the QETPS method with the  BERT-based paragraph selector used in DFGN model~\citep{Xiao:19}. The results of the ablation experiment are shown in the table ~\ref{wo_qetps_table}.

In the table ~\ref{wo_qetps_table}, we find that without using any paragraph filtering method, the Reader is likely to be unable to find the answer to the sub-question from messy paragraphs, resulting in a significant performance degradation. Compared with the results of using the BERT-based paragraph selector in the DFGN model~\citep{Xiao:19}, our QETPS method has achieved better performance, which may be due to our QETPS method makes good use of the entity information in the paragraph, which is just the hop intermediary in multi-hop QA tasks.

\subsection{Results of Evidence Path Generation : Manual Evaluation}
\label{people_test}
Previously in the table ~\ref{main_result_table}, we found that the SRERC model did not perform well in the evidence path metric. However, we analyzed that the unsatisfactory performance is due to that the evidences in the 2WikiMultiHopQA dataset are derived from the tags of the Wikidata knowledge base, which may not appear in the text of question and context. Our SRERC model uses the fragments in the question as the relation in the evidence path, which results in lower score on objective indicators. 

We believe that the evidences of the multi-hop QA model should be expressed in free style, which is difficult to evaluate with objective indicators. As the result, we re-evaluated it through manual evaluation. We randomly selected 100 samples from every question-types to show the evidence path and final predictions of the CRERC model and the SRERC model\footnote{Some cases are shown in the appendix ~\ref{appendix_cases}.}. Each samples was scored by seven graduate students for the evidence extraction capabilities of the two models. We use a score of 1 to 5 to indicate whether the worker believes that the model faithfully demonstrated its reasoning process and got the correct answer. The table ~\ref{people_score_table} shows the results of manual evaluation. We can surprisingly discover that the SRERC model has obtained a higher manual score than the CRERC model. We guess the reason that the expression from the question fragment is easier to reveal the reasoning process of the model. Of course, our conclusions may be biased due to the bias of workers. Therefore, we will continue to explore more rigorous evaluation method for evidence path in our future work.

\section{Conclusion and future work}
\label{conclusion}
We propose a three-stage framework of Relation Extractor-Reader and Comparator (RERC), which solves the multi-hop QA task through the idea of complex question decomposition, and obtains the state-of-the-art results in the 2WikiMultiHopQA dataset, which is close to human performance. Our RERC framework can also provide faithful evidence with excellent interpretability.

Multiple future research directions according to our proposed RERC model may be envisioned. First of all, benefiting to the three-stage structure, the RERC model has the potential to leverage the network structure of the Relation Extractor to gain generalization capabilities for more complex questions. Moreover, we expect that collaborative error correction mechanism applied in Reader module will largely avoid accumulation of errors in multi-hop reasoning.

\section*{Acknowledgements}

The work is supported by The Youth Innovation Promotion Association of the Chinese Academy of Sciences (E1291902), Jun Zhou (2021025). We would like to thank Jiahao Yang, Ming Zhang, Jianzhong Kuang and Chengzhang Li for their valuable support in the procedure of Manual Evaluation. We thank the responsible reviewers for their insightful feedback and valuable suggestions.

\bibliography{anthology,custom}
\bibliographystyle{acl_natbib}

\section*{Appendix}
\appendix

\section{Alternative Methods in Each Stage}
\label{appendix_alter}
In this section, We evaluated several methods in each stage according to the task characteristics which is briefly mentioned in Section ~\ref{method} due to page limit and chose the best one. We show the performance comparison and analysis of those alternative methods in table ~\ref{alter_re} and table ~\ref{alter_reader}.

\begin{table}[htbp]
  \centering
  \resizebox{\columnwidth}{!}{
    \begin{tabular}{lcccc}
    \toprule
    \multicolumn{1}{c}{\multirow{2}[4]{*}{Model}} & \multicolumn{1}{c}{Subject} & \multicolumn{2}{c}{Relations} & \multicolumn{1}{c}{Type} \\
\cmidrule{2-5}          & \multicolumn{1}{c}{F1} & \multicolumn{1}{c}{Acc.} & \multicolumn{1}{c}{F1} & \multicolumn{1}{c}{Acc.} \\
    \midrule
    BERT  & 0.947  & 0.995  & 0.976  & 1.000  \\
    BERT+Transformer & 0.955  & 0.994  & 0.986  & 1.000  \\
    BERT+Type Fuse & 0.947  & 0.999  & 0.994  & 1.000  \\
    BERT+Transformer+Type Fuse & \textbf{0.955 } & \textbf{0.999 } & \textbf{0.997 } & \textbf{1.000 }   \\
    \bottomrule
    \end{tabular}}%
  \caption{The evaluation results of alternative methods for Relation Extractor module. Note that in the Relations area the Acc. is for CRERC model and the F1 value is for SRERC model.}
  \label{alter_re}%
\end{table}%

\begin{table}[htbp]
  \centering
  \resizebox{\columnwidth}{!}{
    \begin{tabular}{ccccc}
    \toprule
    \multirow{2}[4]{*}{\textbf{Model}} & \multicolumn{2}{c}{\textbf{CRERC}} & \multicolumn{2}{c}{\textbf{SRERC}} \\
\cmidrule{2-5}          & EM    & F1    & EM    & F1 \\
    \midrule
    BiDAF & 0.679  & 0.713  & 0.661  & 0.709  \\
    BERT base & 0.835  & 0.862  & 0.803  & 0.841  \\
    BERT large & 0.867  & 0.895  & 0.832  & 0.846  \\
    Roberta base & 0.916  & 0.930  & 0.884  & 0.924  \\
    Roberta large & 0.922  & 0.944  & 0.895  & 0.921  \\
    ALBERT large & \textbf{0.940 } & \textbf{0.958 } & 0.916  & \textbf{0.942 } \\
    ALBERT xlarge & 0.932  & 0.952  & \textbf{0.920 } & 0.932  \\
    \bottomrule
    \end{tabular}}%
  \caption{The evaluation results of alternative methods for Reader module.}
  \label{alter_reader}%
\end{table}%

\section{Output Cases}
\label{appendix_cases}
In this section, we show some cases of CRERC model and SRERC model output for the 2WikiMultiHopQA dataset, including question text, gold label, RE module output (type, subject, relation) and final output (evidence path, prediction) .
\begin{itemize}

\item[$\bullet$] \textbf{Case 1}

\textbf{ID}: 161093c40bde11eba7f7acde48001122

\textbf{Question}: What is the place of birth of Kévin Ledanois's father?

\textbf{Gold Label}: Montreuil

\textbf{Type}: compositional

\textbf{Subject}: Kévin Ledanois

\textbf{-------CRERC Predict-------}

\textbf{Relation}: father; place of birth

\textbf{Evidence}: (Kévin Ledanois; father; Yvon Ledanois)  $\Rightarrow$  (Yvon Ledanois; place of birth; Montreuil-sous-Bois))

\textbf{Predict}: Montreuil-sous-Bois)

\textbf{-------SRERC Predict-------}

\textbf{Relation}: father; place of birth

\textbf{Evidence}: (Kévin Ledanois; father; Yvon Ledanois)  $\Rightarrow$  (Yvon Ledanois; place of birth; Montreuil-sous-Bois))

\textbf{Predict}: Montreuil-sous-Bois)

\hspace*{\fill}

\item[$\bullet$] \textbf{Case 2}

\textbf{ID}: 17ba791a0bde11eba7f7acde48001122

\textbf{Question}: What nationality is the director of film Top Floor Girl?

\textbf{Gold Label}: French

\textbf{Type}: compositional

\textbf{Subject}: Top Floor Girl

\textbf{-------CRERC Predict-------}

\textbf{Relation}: director; country of citizenship

\textbf{Evidence}: (Top Floor Girl; director; Max Varnel)  $\Rightarrow$  (Max Varnel; country of citizenship; French-born)

\textbf{Predict}: French-born

\textbf{-------SRERC Predict-------}

\textbf{Relation}: director; nationality

\textbf{Evidence}: (Top Floor Girl; director; Max Varnel)  $\Rightarrow$  (Max Varnel; nationality; French-born)

\textbf{Predict}: French-born

\hspace*{\fill}

\item[$\bullet$] \textbf{Case 3}

\textbf{ID}: 8f038cdb096011ebbdafac1f6bf848b6

\textbf{Question}: Which film came out earlier, Aram + Aram = Kinnaram or Thayagam?

\textbf{Gold Label}: Aram + Aram = Kinnaram

\textbf{Type}: comparison

\textbf{Subject}: Aram + Aram = Kinnaram; Thayagam

\textbf{-------CRERC Predict-------}

\textbf{Relation}: publication date

\textbf{Evidence}: (Aram + Aram = Kinnaram; publication date; 1985)

(Thayagam; publication date; 1996)

\textbf{Predict}: Aram + Aram = Kinnaram

\textbf{-------SRERC Predict-------}

\textbf{Relation}: came out

\textbf{Evidence}: (Aram + Aram = Kinnaram; came out; 1985)

(Thayagam; came out; 1996)

\textbf{Predict}: Aram + Aram = Kinnaram

\hspace*{\fill}

\item[$\bullet$] \textbf{Case 4}

\textbf{ID}: 17e3349208df11ebbd9fac1f6bf848b6

\textbf{Question}: Who is younger, Osita Chidoka or David Faurschou?

\textbf{Gold Label}: Osita Chidoka

\textbf{Type}: comparison

\textbf{Subject}: David Faurschou; Osita Chidoka

\textbf{-------CRERC Predict-------}

\textbf{Relation}: date of birth

\textbf{Evidence}: (David Faurschou; date of birth; January 28, 1956))

(Osita Chidoka; date of birth; 18 July 1971))

\textbf{Predict}: Osita Chidoka

\textbf{-------SRERC Predict-------}

\textbf{Relation}: younger

\textbf{Evidence}: (David Faurschou; younger; January 28, 1956))

(Osita Chidoka; younger; 18 July 1971))

\textbf{Predict}: Osita Chidoka

\hspace*{\fill}

\item[$\bullet$] \textbf{Case 5}

\textbf{ID}: 8762e83a0baf11ebab90acde48001122

\textbf{Question}: Who is the paternal grandfather of Kerry Earnhardt?

\textbf{Gold Label}: Ralph Earnhardt

\textbf{Type}: inference

\textbf{Subject}: Kerry Earnhardt

\textbf{-------CRERC Predict-------}

\textbf{Relation}: father; father

\textbf{Evidence}: (Kerry Earnhardt; father; Dale Earnhardt)  $\Rightarrow$  (Dale Earnhardt; father; Ralph Earnhardt)

\textbf{Predict}: Ralph Earnhardt

\textbf{-------SRERC Predict-------}

\textbf{Relation}: grandfather; grandfather

\textbf{Evidence}: (Kerry Earnhardt; grandfather; Dale Earnhardt)  $\Rightarrow$  (Dale Earnhardt; grandfather; Ralph Earnhardt)

\textbf{Predict}: Ralph Earnhardt

\hspace*{\fill}

\item[$\bullet$] \textbf{Case 6}

\textbf{ID}: 6a0a17b80baf11ebab90acde48001122

\textbf{Question}: Who is Alice Claypoole Vanderbilt's mother-in-law?

\textbf{Gold Label}: Maria Louisa Kissam

\textbf{Type}: inference

\textbf{Subject}: Alice Claypoole Vanderbilt

\textbf{-------CRERC Predict-------}

\textbf{Relation}: spouse; mother

\textbf{Evidence}: (Alice Claypoole Vanderbilt; spouse; Cornelius Vanderbilt II)  $\Rightarrow$  (Cornelius Vanderbilt II; mother; Maria Louisa Kissam.)

\textbf{Predict}: Maria Louisa Kissam.

\textbf{-------SRERC Predict-------}

\textbf{Relation}: [CLS]; mother

\textbf{Evidence}: (Alice Claypoole Vanderbilt; [CLS]; Cornelius Vanderbilt II)  $\Rightarrow$  (Cornelius Vanderbilt II; mother; Maria Louisa Kissam.)

\textbf{Predict}: Maria Louisa Kissam.

\hspace*{\fill}

\item[$\bullet$] \textbf{Case 7}

\textbf{ID}: 6bc3222c086511ebbd5eac1f6bf848b6

\textbf{Question}: Which film has the director who is older, The Woman Next Door or La Estatua De Carne?

\textbf{Gold Label}: La Estatua De Carne

\textbf{Type}: bridge comparison

\textbf{Subject}: La estatua de carne; The Woman Next Door

\textbf{-------CRERC Predict-------}

\textbf{Relation}: director; date of birth

\textbf{Evidence}: (La estatua de carne; director; Chano Urueta)  $\Rightarrow$  (Chano Urueta; date of birth; February 24, 1904)

(The Woman Next Door; director; François Truffaut)  $\Rightarrow$  (François Truffaut; date of birth; (6 February 1932)

\textbf{Predict}: La estatua de carne

\textbf{-------SRERC Predict-------}

\textbf{Relation}: director; older

\textbf{Evidence}: (La estatua de carne; director; Chano Urueta)  $\Rightarrow$  (Chano Urueta; older; February 24, 1904)

(The Woman Next Door; director; François Truffaut)  $\Rightarrow$  (François Truffaut; older; (6 February 1932)

\textbf{Predict}: La estatua de carne

\hspace*{\fill}

\item[$\bullet$] \textbf{Case 8}

\textbf{ID}: 09646113087011ebbd62ac1f6bf848b6

\textbf{Question}: Which film has the director died later, Fugitives For A Night or Chinese In Paris?

\textbf{Gold Label}: Chinese In Paris

\textbf{Type}: bridge comparison

\textbf{Subject}: Fugitives for a Night; Chinese in Paris

\textbf{-------CRERC Predict-------}

\textbf{Relation}: director; date of death

\textbf{Evidence}: (Fugitives for a Night; director; Leslie Goodwins)  $\Rightarrow$  (Leslie Goodwins; date of death; 8 January 1969))

(Chinese in Paris; director; Jean Yanne)  $\Rightarrow$  (Jean Yanne; date of death; 23 May 2003))

\textbf{Predict}: Chinese in Paris

\textbf{-------SRERC Predict-------}

\textbf{Relation}: director; die

\textbf{Evidence}: 

(Fugitives for a Night; director; Leslie Goodwins)  $\Rightarrow$  (Leslie Goodwins; die; 8 January 1969))

(Chinese in Paris; director; Jean Yanne)  $\Rightarrow$  (Jean Yanne; die; 23 May 2003))

(8 January 1969; less than; 23 May 2003)

\textbf{Predict}: Chinese in Paris
\end{itemize}

\end{document}